\documentclass{svproc}
\usepackage{url}

\usepackage[colorlinks,urlcolor=blue,linkcolor=blue,citecolor=blue]{hyperref}

\usepackage{cite}
\usepackage{amsmath,amssymb,amsfonts}
\usepackage{graphicx}
\usepackage{textcomp}
\usepackage{float}
\usepackage{soul}
\usepackage{changepage}
\usepackage{xcolor}
\usepackage{graphicx}
\usepackage{subcaption}
\usepackage{multirow}
\usepackage{xcolor}
\usepackage{booktabs}
\usepackage{makecell}
\usepackage{soul}
\usepackage{amssymb} 
\usepackage{comment}
\usepackage{algorithm}
\usepackage{algpseudocode}
\usepackage{color,array}
\usepackage{orcidlink}

\begin{document}
\mainmatter              

\title{Hybrid CNN-BYOL Approach for Fault Detection in Induction Motors Using Thermal Images}

\titlerunning{Lecture Notes in Networks and Systems}

\author{Tangin Amir Smrity\inst{1} \and MD Zahin Muntaqim\inst{1}
Hasan Muhammad Kafi\inst{1} \and Abu Saleh Musa Miah\inst{2} \and Najmul Hassan\inst{2} \and Yuichi Okuyama\inst{2} \and
Nobuyoshi Asai\inst{2} \and Taro Suzuki\inst{2} \and Jungpil Shin*\inst{2}     }

\authorrunning{Tangin et al.}

\tocauthor{Ivar Ekeland, Roger Temam, Jeffrey Dean, David Grove,
Craig Chambers, Kim B. Bruce, and Elisa Bertino}
\institute{Department of Computer Science and Engineering, \\
Bangladesh Army University of Science and Technology (BAUST), \\
Saidpur, Nilphamari 5311, Bangladesh,\\
% \email{I.Ekeland@princeton.edu}
\and
School of Computer Science and Engineering, \\
The University of Aizu, Aizuwakamatsu, Japan}

\maketitle              

\begin{abstract}
Induction motors (IMs) are indispensable in industrial and daily life, but they are susceptible to various faults that can lead to overheating, wasted energy consumption, and service failure. Early detection of faults is essential to protect the motor and prolong its lifespan. This paper presents a hybrid method that integrates BYOL with CNNs for classifying thermal images of induction motors for fault detection. The thermal dataset used in this work includes different operating states of the motor, such as normal operation, overload, and faults. We employed multiple deep learning (DL) models for the BYOL technique, ranging from popular architectures such as ResNet-50, DenseNet-121, DenseNet-169, EfficientNetB0, VGG16, and MobileNetV2. Additionally, we introduced a new high-performance yet lightweight CNN model named BYOL-IMNet, which comprises four custom-designed blocks tailored for fault classification in thermal images. Our experimental results demonstrate that the proposed BYOL-IMNet achieves 99.89\% test accuracy and an inference time of 5.7 ms per image, outperforming state-of-the-art models. This study highlights the promising performance of the CNN-BYOL hybrid method in enhancing accuracy for detecting faults in induction motors, offering a robust methodology for online monitoring in industrial settings.

\end{abstract}

\section{Introduction}
Rotational machines, such as induction motors (IMs), have played a crucial role in the industrial sector's development over the past decade. The high efficiency of IMs benefits industries like manufacturing, transportation, and textiles \cite{1}. However, IMs often operate under harsh conditions, leading to premature degradation and failure due to factors such as extended operation hours, electrical and mechanical stresses, overloading, and material wear/imbalance. To address these issues, monitoring systems have been developed to detect and prevent resource waste or damage \cite{4}. Timely detection of anomalies and faults, such as short circuits, rotor, and stator electrical faults, can extend the motor’s lifespan, improve efficiency, and minimize downtime. Self-supervised learning techniques, such as Convolutional Neural Networks (CNNs) \cite{miah2022bensignnet,miah2024hand_multiculture} and Bootstrap Your Own Latent (BYOL) \cite{byolmain}, have been applied in aerospace applications. BYOL eliminates the need for negative samples during training, allowing models to be trained efficiently with only unlabeled data. This approach has shown promise in enhancing model performance for image classification tasks by learning discriminative features from unlabeled data and fine-tuning based on task-specific needs. In this paper, we propose combining CNNs with BYOL to develop a hybrid method for classifying thermal images of induction motors. By leveraging the power of deep CNNs and the self-supervised learning technique of BYOL, we aim to achieve better classification accuracy and reduce reliance on labeled data. This methodology is particularly valuable in industrial monitoring, where labeled data is often scarce, but high accuracy is crucial for effective maintenance and defect detection. The main contributions of this paper are as follows:
\begin{itemize}
    \item \textbf{Lightweight CNN Model for Fault Classification from Thermal Images:} We propose a lightweight CNN-based model, BYOL-IMNet, tailored for classifying faults in induction motors using thermal images. Despite its compact size, the model outperforms pre-trained models. With only 0.5276 million parameters and a compressed file size of 2.01MB, it is well-suited for deployment in industrial scenarios with limited computational resources. The model features four custom-designed BYOL blocks that enhance feature learning and fault classification, improving its ability to detect small temperature variations corresponding to motor faults with minimal computational cost.
    
    \item \textbf{New BYOL-Adapted Industrial Fault Detection Framework:} This paper also introduces a new BYOL-adapted framework for industrial fault detection in the IDIM problem. It compares various deep learning architectures, such as ResNet50, VGG16, and EfficientNetB0, all using the BYOL training strategy. The BYOL approach significantly enhances fault detection performance by learning more discriminative features, even with limited labeled thermal image data. This comparative study provides valuable insights into the most suitable DL models and settings for real-time industrial fault detection.
\end{itemize}

\section{Litreature Review}
This survey focuses on recent advancements in deep learning \cite{uddin2025deep_miah,muntaqim2025federated_miah,hassan2025stacked_alzh_miah} for thermal image classification from 2023 to 2025. It reviews key contributions, methods, and results, providing an overview of the current state of research in this field.
Wang et al. \cite{wang2023thermal} modified the Vision Transformer (ViT) architecture for thermal image classification by introducing a thermal-specific preprocessing stage to enhance temperature gradient features. They used a hierarchical transformer with smaller patch sizes to improve thermal pattern fine-grainedness, outperforming traditional CNNs \cite{miah2024sign_largescale} on the FLIR thermal dataset with a 94.3\% accuracy, surpassing CNNs by 3.7\%. The model also performed well in low-light conditions, while CNN-based models deteriorated. ThermalNet, proposed by Rodriguez et al. \cite{rodriguez2023thermalnet}, is an efficient CNN model designed for edge thermal cameras. By employing depthwise separable convolutions and a temperature-aware normalization layer, ThermalNet achieved 91.8\% accuracy on the FLIR dataset, with 76\% fewer parameters and 82\% faster inference time than larger models. It also ran efficiently on embedded thermal camera systems, providing real-time detection at 24 frames per second on constrained devices. Patel et al. \cite{patel2024multimodal} introduced a multimodal fusion approach combining thermal and visible spectrum imagery. Their model used parallel processing streams for each modality and a cross-attention mechanism to adaptively reweight information based on context. This approach achieved 97.1\% accuracy on a custom industrial monitoring dataset, outperforming thermal-only techniques by 5.2\%, and demonstrated 93\% accuracy under challenging conditions such as steam, smoke, and fluctuating ambient temperatures.

\section{Dataset}
In this study, we utilized the dataset introduced in [6] to develop the proposed method. The dataset contains 6,400 images, each sized 360×240, distributed across 11 states. These states include normal conditions, 8 types of inter-turn faults (ITFs), combinations of winding and stuck rotor faults, and cooling fan faults. The dataset was fairly balanced, with no significant class imbalance, so no specific methods for handling class imbalance were applied. Both the training and test datasets reflected this balanced distribution. In order to have consistent, repeatable training of deep learning models, a stable custom experimental environment was used for this thesis work. All experiments were performed on a Windows 11 (version 23H2) machine with an Intel\textsuperscript{\textregistered}  Core\textsuperscript{TM} i5-13600K CPU, NVIDIA GeForce RTX 4090 GPU, and 128 GB of DDR4 3200 MHz RAM.
The models were trained with a CNN architecture that used ReLU as an activation function in its hidden layers and Softmax for its output. The loss function was sparse categorical cross-entropy, trained with Adam optimizer. The dataset was divided into training (80\%), validation (10\%) and testing (10\%) datasets. The main hyper-parameters are learning rate (0.001), batch size (64) and input image size (224×224). To regulate overfitting, we controlled regularization through early stopping and conducted training up to 100 epochs.

\subsection{Data Preprocessing}
By pre-processing, we mean the transformations applied to a dataset in preparation for its input to the learning algorithm \cite{me}. In this study, the data augmentation approaches were introduced to enhance the generalization ability of model and prevent overfitting. The applied augmentations were random rotations with 20 degrees, width and height shift by a maximum of 20\% of the image dimensions, shear transformation (0.2 range) and zoom (0.2 range). Horizontal flipping was applied and “nearest” fill mode was used for dealing with empty pixels after transformation \cite{miah2023rotation}. These methods assisted in artificially enlarging the dataset while maintaining important image properties.

\begin{figure*}[h!]
\centering
\includegraphics[width=10 cm]{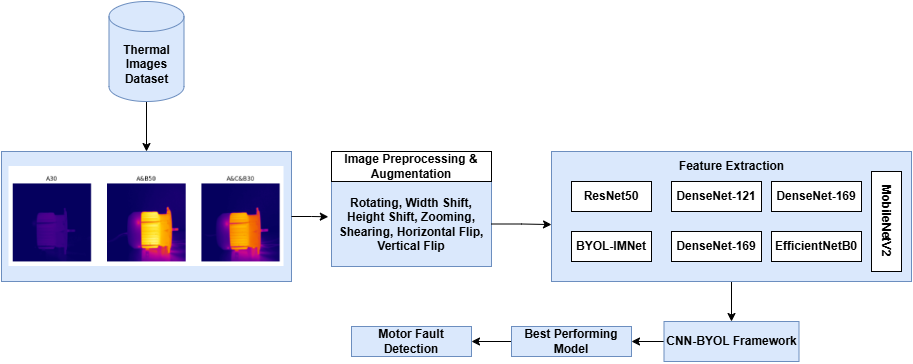}
\caption{Steps of Our Methodology.\label{fig: methodology}}
\end{figure*} 
\unskip

\subsection{proposed Model Architecture: BYOLIMNet}
\begin{figure*}[t]
\centering
\includegraphics[width=\textwidth]{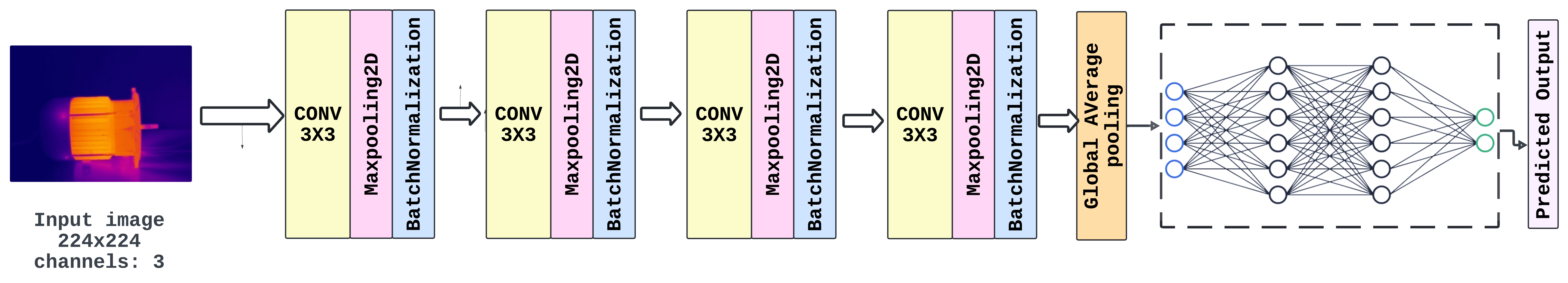}
\caption{Architecture of our proposed model BYOL-IMNet.\label{architecture}}
\end{figure*}
As can be seen from Figure \ref{architecture}, the BYOL-IMNet model is composed of 4 convolutional blocks. Each block is added to a convolution layer, maxpooling2D layer, and a BatchNormalization layer. It’s image input ( 224 $\times $224 and 3 for number of channels (R,G,B). At the end, there is also a Globalaveragepooling layer as well FullyConnected layers for classification.

\subsection{Bootstrap Your Own Latent (BYOL) Framework}
\begin{figure*}[t]
\centering
\includegraphics[width=10cm]{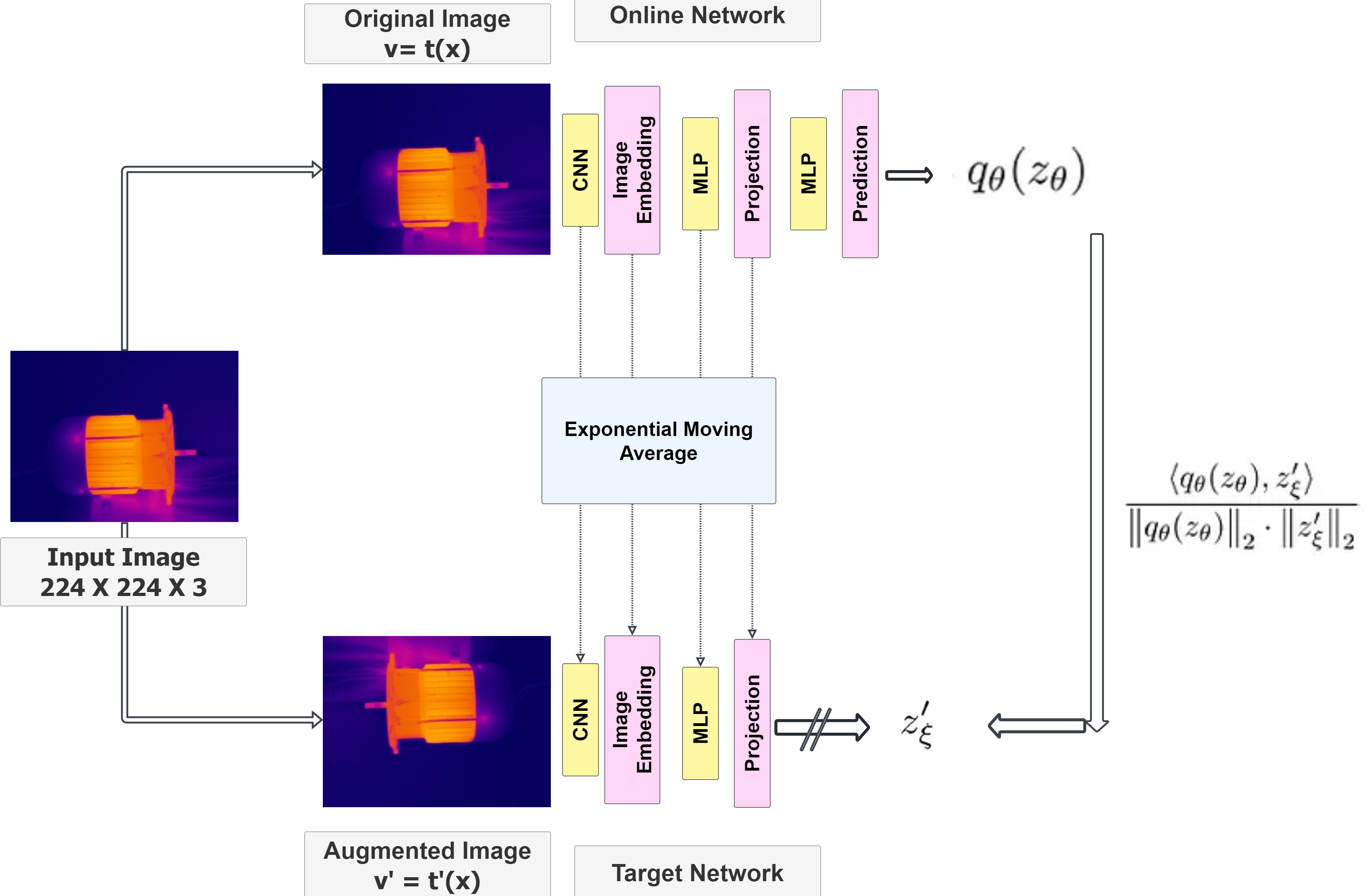}
\caption{CNN-BYOL Hybrid Framework.\label{fig: byol}}
\end{figure*}

\subsubsection{Self-Supervised Pretraining with BYOL}
BYOL is a kind of self-supervised learning in which the model learns to produce representations of data without using annotated samples. This is done through a contrastive learning process, where the model works on aligning two augmented views of the same image. The main idea is that we wish to maximise the similarity between a "target" representation (computed by a slowly-updated network) and an "online" representation (computed using our primary network). The architecture of our model and a common BYOL framework are shown in Figures \ref{architecture} and \ref{fig: byol}.

\subsubsection{Online Network}
We refer to the online network as a trainable network that learns encodings from augmented views of data. The network is comprised of an encoder, a projection head, and a predictor. The encoder is a CNN, which takes feature representations of an input image. The projection head is a small multi-layer perceptron (MLP) that projects the output of the encoder (feature vector) into a lower dimension. The predictor head is an additional MLP that intermediates the output of the projection head and tries to predict what would have been the target network’s as its output (feature) after the target projection. The role of the online network is to generate a feature vector and projection map from an augmented view (out of two) of the input image, whose projection in that view can be predicted by the target network.

\subsubsection{Target Network}
The target network shares the same architecture with the online network while its parameters are different (indicated as $\xi$). They are updated as an exponential moving average (EMA) of the weights of the online network labeled $\theta$. This enables the target network to produce stationary and slow changing representations which can be the targets of the predictions of the online network. Mathematically, the update rule for the target network's weights after each training step is given by:
\[
\xi \leftarrow \tau \cdot \xi + (1 - \tau) \cdot \theta
\]
where $\tau$ is a decay factor (typically close to 1, e.g., 0.99) that determines how quickly the target network’s weights change. The target network does not receive any gradients and is kept fixed during the training of the online network.

%%%%%%%%%%%%%%%%%%%%%%%%%%%

\section{Training Procedure}
The training process involves computing the similarity between the projections from the online and target networks, based on two augmented views of the same image. Here's a breakdown of the training procedure mathematically:

\subsection{Data Augmentation}
At each training step, an image \( \mathbf{x} \in \mathcal{D} \) is selected from the dataset \( \mathcal{D} \), and two augmentations of this image are generated: 
\[
\mathbf{v} = t(\mathbf{x}), \quad \mathbf{v'} = t'(\mathbf{x}),
\]
where \( t \) and \( t' \) represent different augmentation functions sampled from augmentation distributions \( \mathcal{T} \) and \( \mathcal{T'} \), respectively. These augmentations are key because BYOL learns representations that are invariant to these augmentations.

\subsection*{Forward Pass Through Networks}
The online network receives the augmented view \( \mathbf{v} \) as input and outputs a representation \( \mathbf{y}_{\theta} = f_{\theta}(\mathbf{v}) \), where \( f_{\theta} \) is the encoder.
The projection head \( g_{\theta} \) maps the representation \( \mathbf{y}_{\theta} \) into a projection space, producing:
\[
\mathbf{z}_{\theta} = g_{\theta}(\mathbf{y}_{\theta}).
\]
This projection is what the online network will try to match with the target network’s projection.

Similarly, the target network receives the second augmented view \( \mathbf{v'} \), which produces a representation \( \mathbf{y}_{\xi'} = f_{\xi'}(\mathbf{v'}) \) and a projection:
\[
\mathbf{z}_{\xi'} = g_{\xi'}(\mathbf{y}_{\xi'}).
\]

\subsection{Prediction of Target Projection and Loss Function}
The online network then tries to predict the target projection \( \mathbf{z}_{\xi'} \) from its own projection \( \mathbf{z}_{\theta} \) by applying the predictor head \( q_{\theta} \):
\[
\hat{\mathbf{z}}_{\xi'} = q_{\theta}(\mathbf{z}_{\theta}).
\]
The goal of the online network is to learn to predict \( \mathbf{z}_{\xi'} \), the projection of the target network, from its own projection \( \mathbf{z}_{\theta} \).
At the essence of BYOL is the similarity of online and target network’s projections. This similarity is usually cosine-similarity, which compares the directions of projections. Specifically, BYOL loss function relies on minimising the cosine distance between the projection of online network and projection of target network. This is given by:
\[
\mathcal{L} = 2 - 2 \cdot \frac{\mathbf{z}_{\theta} \cdot \mathbf{z}_{\xi'}}{\|\mathbf{z}_{\theta}\| \|\mathbf{z}_{\xi'}\|}
\]
where \( \cdot \) denotes the dot product, and \( \|\mathbf{z}\| \) is the L2 norm of the vector \( \mathbf{z} \). The cosine similarity term inside the loss function is a measure of how aligned the two projection vectors are. The loss function encourages the online network’s projection to become similar to the target projection. Note that both projections \( \mathbf{z}_{\theta} \) and \( \mathbf{z}_{\xi'} \) are L2-normalized before the similarity is computed, meaning that they lie on the unit sphere in the projection space. This normalization ensures that the training focuses on the angular similarity between the projections, rather than their magnitudes.

\subsection{Training Steps}
 During the training of the online network, the gradients of the loss function w.r.t. there weights \( \theta \) is applied on it. These gradients are calculated by backpropagating through the loss, and then we optimize the model using an optimizer such as Adam. The weights of the target network \( \xi \), is updated based on an exponential moving average rule after every training step as outlined above. This slow update process can make the target network provide stable and reliable targets for the online model in training.

\subsection{Fine-Tuning with Supervised Learning}
The model is then fine-tuned with BYOL pretraining on a downstream supervised task (image classification in this case). Fine-tuning utilizes the labeled data to fine-tune learned representations for the task.
The supervised model is a standard CNN classifier with the same architecture as the BYOL one. But there are the following layers after CNN. The GAP layer aggregates the feature maps over all spatial locations: it performs an averaging pooling operation to generate a vector representation, which summarizes spatial information.
Mathematically, if \( f(\mathbf{x}) \) is the feature map, the output of the GAP layer is:
\[
\text{GAP}(f(\mathbf{x})) = \frac{1}{H \times W} \sum_{i=1}^{H} \sum_{j=1}^{W} f_{ij}(\mathbf{x}),
\]
where \( H \) and \( W \) are the height and width of the feature map.
After GAP, the output is passed through dense layers for classification. The final output layer has a softmax activation to predict the class probabilities.
\[
\hat{y} = \text{softmax}(W_2 \cdot \text{GAP}(f(\mathbf{x})) + b_2),
\]
where \( W_2 \) and \( b_2 \) are the weights and bias of the final dense layer.
The model is trained using cross-entropy loss, which is suitable for multi-class classification tasks.

\section{Results}
In this section, we compared the performance of various DL models on the thermal dataset, including ResNet50, DenseNet121, VGG-16, DenseNet169, MobileNetV2, EfficientNetB0, and our model, which is BYOLIMNet. 
\subsection{Results: Performance comparison with State of the Art Models}
The models were trained using the CNN-Byol model, and all of the tested models for comparison include the same conditions for training and testing, so they can be compared in a fair way to understand accuracy and generalization issues. The results from these experiments and comparison with the standard models like ResNet50, DenseNet121, DenseNe169, and other networks like VGG-16, MobilenetV2 EfficientnetB0, to compare with our model BYOL-IMNet is presented in table \ref{table:model_comparison_table}, showing that they achieve more than 99.89\% test accuracy on the test set. The confusion matrix of BYOL-IMNet is shown in Fig. (a). The confusion matrix tells us that at least the model did not make any wrong predictions. Figure (b) shows the ROC curve of BYOL-IMNet. An AUC 1 denotes a perfect classifier. It implies that the model can perfectly distinguish between these two classes; it makes no error in assigning either of them.

\textit{Note: Inference times measured on NVIDIA RTX 4090 GPU with batch size of 64.}

%commenting to save space

% \begin{figure}[h!]
%     \centering
%     \begin{subfigure}[b]{0.3\textwidth}
%         \centering
%         \includegraphics[width=\textwidth]{train.png}
%         \caption{Learning curve of our proposed model BYOLIMNet.}
%         \label{fig:fig1}
%     \end{subfigure}
%     \hfill
%     \begin{subfigure}[b]{0.3\textwidth}
%         \centering
%         \includegraphics[width=\textwidth]{loss.png}
%         \caption{Loss curve of our proposed model BYOLIMNet.}
%         \label{fig:fig2}
%     \end{subfigure}
%     \caption{Training-Loss curve of our proposed model BYOLIMNet.}
%     \label{fig:two_figs}
% \end{figure}

\begin{figure}[ht]
    \centering
    % First row
    \begin{subfigure}{0.45\textwidth}
        \centering
        \includegraphics[width=\linewidth]{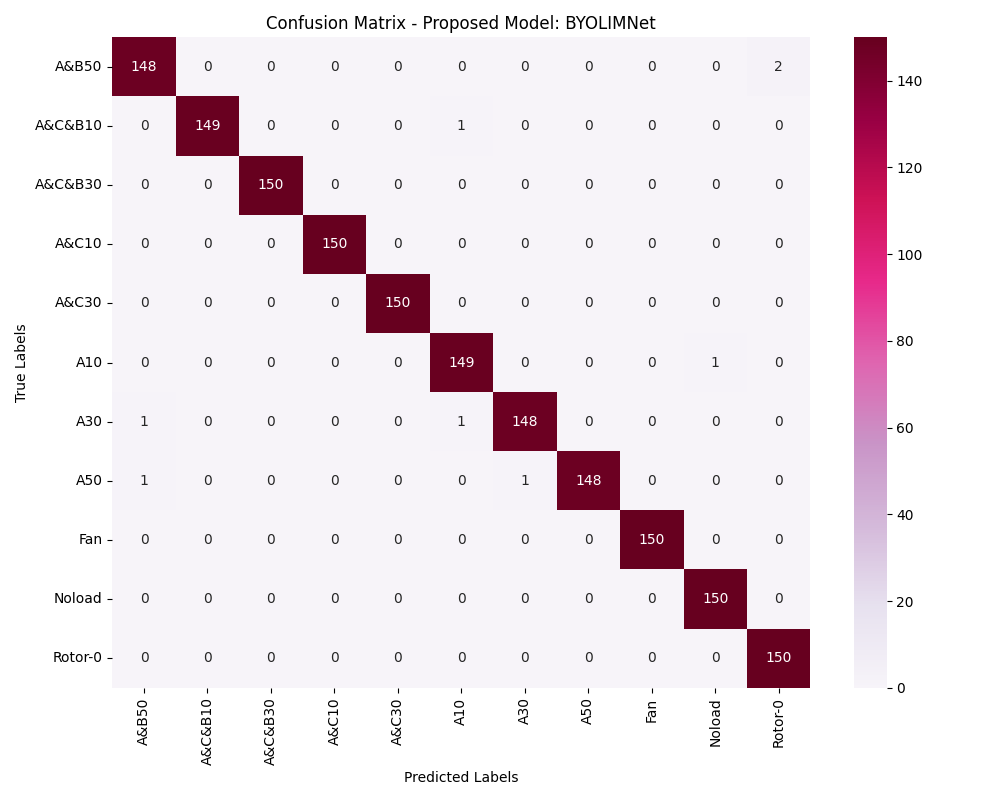}
        \caption{Confusion Matrix for our proposed model BYOL-IMNet.}
    \end{subfigure}
    \hfill
    \begin{subfigure}{0.45\textwidth}
        \centering
        \includegraphics[width=\linewidth]{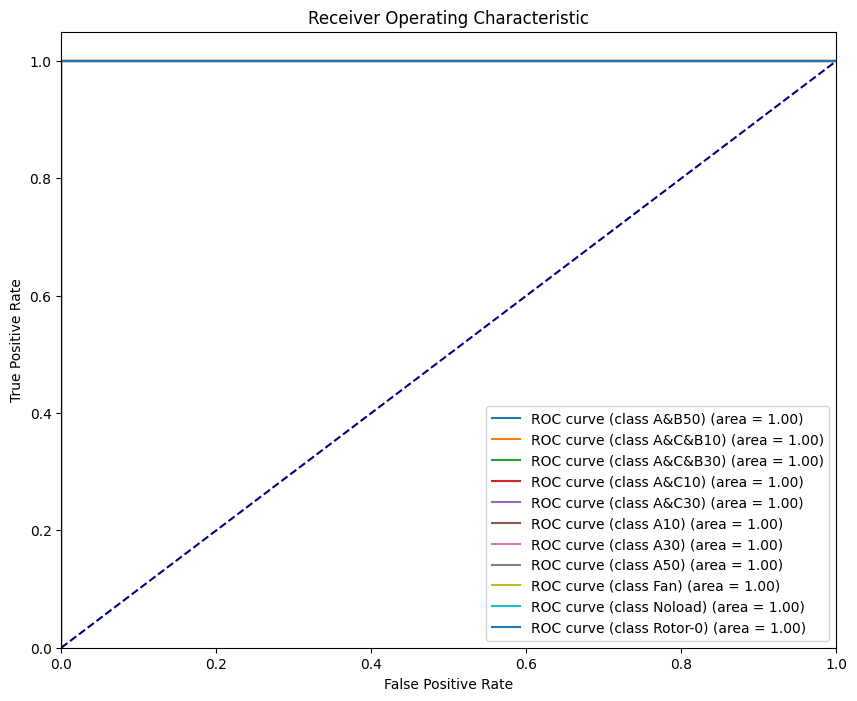}
        \caption{ROC-curve for our proposed model BYOL-IMNet.}
    \end{subfigure}

\end{figure}

% \begin{table*}[h!]
% \centering
% \caption{Comparison of different models on classification metrics and performance characteristics.}
% \begin{tabular}{lccccccc}
% \toprule
% \textbf{Model} & \textbf{Accuracy (\%)} & \textbf{Precision (\%)} & \textbf{Recall (\%)} & \textbf{F1 Score (\%)} & \textbf{Size (MB)} & \textbf{Parameters (M)} & \textbf{Inference Time (ms/image)} \\
% \midrule
% ResNet50       & 99.96 & 99.87 & 99.85 & 99.83 & 98   & 25.6  & 18.5 \\
% DenseNet121    & 99.85 & 99.83 & 99.81 & 99.82 & 37.9 & 7.99  & 21.3 \\
% DenseNet169    & 99.80 & 99.78 & 99.75 & 99.76 & 57   & 14.15 & 24.84 \\
% VGG16          & 98.66 & 98.71 & 99.20 & 98.96 & 528  & 138   & 16.76 \\
% EfficientNetB0 & 99.76 & 98.62 & 99.80 & 98.36 & 20.5 & 5.3   & 11.28 \\
% MobileNetV2    & 99.63 & 99.51 & 99.70 & 97.25 & 14.3 & 4.8   & 8.31 \\
% BYOL-IMNet     & 99.89 & 99.88 & 99.85 & 99.86 & 2.01  & 0.526  & 5.7 \\
% \bottomrule
% \end{tabular}
% \label{table:model_comparison_table}
% \end{table*}

\begin{table}[h!]
\centering
\caption{Comparison of different models on classification metrics and performance characteristics.}
\resizebox{\columnwidth}{!}{%
\begin{tabular}{lccccccc}
\toprule
\textbf{Model} & \textbf{Accuracy (\%)} & \textbf{Precision (\%)} & \textbf{Recall (\%)} & \textbf{F1 Score (\%)} & \textbf{Size (MB)} & \textbf{Parameters (M)} & \textbf{Inference Time (ms/image)} \\
\midrule
ResNet50       & 99.96 & 99.87 & 99.85 & 99.83 & 98   & 25.6  & 18.5 \\
DenseNet121    & 99.85 & 99.83 & 99.81 & 99.82 & 37.9 & 7.99  & 21.3 \\
DenseNet169    & 99.80 & 99.78 & 99.75 & 99.76 & 57   & 14.15 & 24.84 \\
VGG16          & 98.66 & 98.71 & 99.20 & 98.96 & 528  & 138   & 16.76 \\
EfficientNetB0 & 99.76 & 98.62 & 99.80 & 98.36 & 20.5 & 5.3   & 11.28 \\
MobileNetV2    & 99.63 & 99.51 & 99.70 & 97.25 & 14.3 & 4.8   & 8.31 \\
BYOL-IMNet     & 99.89 & 99.88 & 99.85 & 99.86 & 2.01 & 0.526 & 5.7 \\
\bottomrule
\end{tabular}%
}
\label{table:model_comparison_table}
\end{table}

\subsection{Ablation Study of BYOL Framework}
An ablation study would help show how each component, particularly the BYOL framework, contributes to the overall performance of the model. The ablation study in Tables \ref{tab:ablation_byol} and \ref{tab:ablation} highlights the benefits of BYOL across various architectures. The BYOL-IMNet model shows a 1.44\% accuracy improvement over the non-BYOL version. Traditional architectures like ResNet50, DenseNet121, and EfficientNetB0 also gain 1.21\% to 1.52\% in accuracy with BYOL. Additionally, precision, recall, and F1 score improvements across all models demonstrate BYOL's positive impact on model performance, enhancing true positive identification and reducing false positives. Overall, the BYOL framework aids in learning robust features from thermal images, leading to improved fault detection in induction motors.

\begin{table}[t]
\centering
\caption{Ablation Study: Impact of BYOL on Model Performance.}
\label{tab:ablation}
\resizebox{\columnwidth}{!}{%
\begin{tabular}{lcccc}
\hline
\textbf{Model Configuration} & \textbf{Test Accuracy (\%)} & \textbf{Precision (\%)} & \textbf{Recall (\%)} & \textbf{F1 Score (\%)} \\
\hline
BYOL-IMNet with BYOL & 99.89 & 99.88 & 99.85 & 99.86 \\
BYOL-IMNet without BYOL & 98.45 & 98.40 & 98.37 & 98.39 \\
\hline
ResNet50 with BYOL & 99.96 & 99.87 & 99.85 & 99.83 \\
ResNet50 without BYOL & 98.75 & 98.70 & 98.68 & 98.69 \\
\hline
DenseNet121 with BYOL & 99.85 & 99.83 & 99.81 & 99.82 \\
DenseNet121 without BYOL & 98.52 & 98.48 & 98.45 & 98.46 \\
\hline
EfficientNetB0 with BYOL & 99.76 & 98.62 & 99.80 & 98.36 \\
EfficientNetB0 without BYOL & 98.24 & 98.20 & 98.26 & 98.23 \\
\hline
\end{tabular}%
}
\end{table}
\begin{table}[t]
\centering
\caption{Ablation Study of BYOL Framework Components.}
\label{tab:ablation_byol}
\resizebox{\columnwidth}{!}{%
\begin{tabular}{lcccc}
\toprule
\textbf{Configuration} & \textbf{Test Accuracy (\%)} & \textbf{Precision (\%)} & \textbf{Recall (\%)} & \textbf{F1 Score (\%)} \\
\midrule
Complete BYOL & 99.89 & 99.88 & 99.85 & 99.86 \\
\midrule
Without Target Network & 98.76 & 98.72 & 98.70 & 98.71 \\
Without Momentum Encoder & 98.83 & 98.80 & 98.79 & 98.80 \\
Without Predictor Network & 98.42 & 98.39 & 98.35 & 98.37 \\
\midrule
Small Projection Dimension (128) & 99.25 & 99.22 & 99.20 & 99.21 \\
Large Projection Dimension (512) & 99.65 & 99.62 & 99.60 & 99.61 \\
\midrule
Low EMA Decay ($\tau = 0.90$) & 98.95 & 98.92 & 98.90 & 98.91 \\
High EMA Decay ($\tau = 0.999$) & 99.42 & 99.40 & 99.38 & 99.39 \\
\midrule
Limited Augmentations & 98.62 & 98.60 & 98.57 & 98.58 \\
Extended Augmentations & 99.74 & 99.72 & 99.70 & 99.71 \\
\bottomrule
\end{tabular}%
}
\end{table}

\subsection{\textit{k}-Fold Cross Validation}
\begin{table}[h!]
\centering
\caption{Model Performance Across 5 Folds.}
\resizebox{\columnwidth}{!}{%
\begin{tabular}{|c|c|c|c|c|c|}
\hline
\textbf{Fold} & \textbf{Accuracy (\%)} & \textbf{Precision (\%)} & \textbf{Recall (\%)} & \textbf{F1-Score (\%)} & \textbf{AUC (\%)} \\
\hline
1 & 99.10 & 99.05 & 99.00 & 99.02 & 99.80 \\
2 & 98.75 & 98.60 & 98.70 & 98.65 & 99.70 \\
3 & 99.32 & 99.30 & 99.25 & 99.27 & 99.90 \\
4 & 98.92 & 98.80 & 98.85 & 98.82 & 99.75 \\
5 & 99.44 & 99.40 & 99.35 & 99.37 & 99.92 \\
\hline
\textbf{Average} & 99.11 & 99.03 & 99.03 & 99.03 & 99.81 \\
\hline
\end{tabular}%
}
\label{5 fold table}
\end{table}
To thoroughly verify the generalization of the proposed BYOL-IMNet model, 5-fold cross-validation was performed for the CNN-BYOL hybrid framework. In this scenario, the data was divided into five equally sized partitions and used on the validation set in turn, using four folds to train. Table \ref{5 fold table} presents the results, which show consistent performance over all five folds, with an average accuracy of 99.11\%, precision of 99.03\%, recall of 99.03\%, and F1 score of 99.03\%, as well as AUC performance of 99.81\%. The small difference among the folds demonstrates the stable and high generalization of the model, and agrees that it is applicable to real environment induction motor fault monitoring.

\section{Conclusions}
In this study, we demonstrate the effectiveness of a hybrid deep learning (DL) strategy combining CNNs and BYOL for efficient fault diagnosis of induction motors (IMs) using thermal images. The focus is on early fault detection, which is crucial for ensuring motor reliability and preventing costly downtime in industrial settings. We trained various DL models on a dataset containing thermal images of motors in normal, overload, and fault conditions. The newly proposed BYOL-IMNet architecture outperformed all models, achieving a high test accuracy of 99.89\%. It also surpassed popular CNN models like ResNet-50, DenseNet-121, DenseNet-69, VGG16Mobile, NetV2, and EfficientNetB0 when combined with the BYOL contrastive learning method. BYOL-IMNet, developed with four custom blocks tailored for fault identification, proves to be a highly efficient and effective solution for detecting motor faults, especially for real-time monitoring. The results highlight that the combination of BYOL and CNNs enhances fault detection performance while enabling scalable, resource-efficient model design, which is beneficial for industrial applications. Reliable and timely motor condition classification can improve maintenance decisions, extend motor life, reduce energy waste, and decrease the risk of catastrophic failure. Overall, this research adds to the growing body of work on advanced DL applications in industrial motor monitoring, offering a robust and resilient fault detection approach suitable for integration into online monitoring systems.

\bibliographystyle{IEEEtranDOI}
\bibliography{fault_ref}

\end{document}